\documentclass{article}

\usepackage[preprint]{neurips_2024}

\usepackage[utf8]{inputenc} 
\usepackage[T1]{fontenc}    
\usepackage{hyperref}       
\usepackage{bookmark}       
\usepackage{url}            
\usepackage{booktabs}       
\usepackage{amsfonts}       
\usepackage{nicefrac}       
\usepackage{microtype}      
\usepackage{xcolor}         
\usepackage{graphicx}
\usepackage{booktabs}
\usepackage{array}
\usepackage{bm}

\title{Trust in Disinformation Narratives: 
a \textit{Trust in the News} Experiment}

\author{%
  Hanbyul Song \\
  Universit\`{e} Lorraine\\
  \texttt{hanbyulsong91@gmail.com} \\
  \And
  Miguel F. Santos Silva \\
  Blanquerna University \\
  \texttt{miguelfd@blanquerna.url.edu} \\
  \AND
  Jaume Suau \\
  Blanquerna University \\
  \texttt{jaumesm@blanquerna.url.edu} \\
  \And
  Luis Espinosa-Anke \\
  CardiffNLP, Cardiff University / AMPLYFI \\
  \texttt{espinosa-ankel@cardiff.ac.uk} \\
}

\makeatletter
\providecommand{\@notice}{}
\makeatother

\begin{document}

\maketitle

\begin{abstract}
Understanding why people trust or distrust one another, institutions, or information is a complex task that has led scholars from various fields of study to employ diverse epistemological and methodological approaches \cite{isaeva2015why}. Despite the challenges, it is generally agreed that the antecedents of trust (and distrust) encompass a multitude of emotional and cognitive factors, including a general disposition to trust and an assessment of trustworthiness factors \cite{baer2018not}. In an era marked by increasing political polarization \cite{mudde2017populism}, cultural backlash \cite{norris2022praise}, widespread disinformation and fake news \cite{oconnor2019misinformation}, and the use of AI software to produce news content \cite{ding2023harnessing}, the need to study trust in the news has gained significant traction.

This study presents the findings of a \textit{trust in the news} experiment designed in collaboration with Spanish and UK journalists, fact-checkers, and the CardiffNLP Natural Language Processing research group. The purpose of this experiment, conducted in June 2023, was to examine the extent to which people trust a set of fake news articles based on previously identified disinformation narratives related to gender, climate change, and COVID-19.

The online experiment participants (801 in Spain and 800 in the UK) were asked to read three fake news items and rate their level of trust on a scale from 1 (not true) to 8 (true). The pieces used a combination of factors, including stance (favourable, neutral, or against the narrative), presence of toxic expressions, clickbait titles, and sources of information to test which elements influenced people's responses the most. Half of the pieces were produced by humans and the other half by ChatGPT.

The results show that the topic of news articles, stance, people’s age, gender, and political ideologies significantly affected their levels of trust in the news, while the authorship (humans or ChatGPT) does not have a significant impact.

\end{abstract}

\section{Introduction}
\label{sec:introduction}

In ``Only the sheep trust journalism?'', argues that most US citizens tend to distrust the news and the work of journalists in general because they find it socially desirable to ``be seen as smart, savvy, confident, and critically distant from ‘biased’ journalists -to not, in effect, be one of those 'sheep' who blindly follow''. Whereas trust in the media and in public institutions is considered paramount to ensure the proper functioning of democracy \cite{vanaelst2017political}, recent studies show that trust in journalism, the media, and public/political institutions is traditionally low \cite{fawzi2021concepts}. According to the latest Edelman Trust Barometer Global Report (2023), various countries undergo increasing levels of distrust in government leaders, journalists, and media outlets as well as growing political polarization. 

Different studies have tried to find correlations between the levels of trust in the media and various major economic, social, and political factors. For instance, \cite{zuniga2019trust} found that countries with lower Human Development Index scores present higher levels of trust in the media. While \cite{tsfati2014individual} observed that countries with advanced economic development levels, that is, with a higher gross domestic product, are likely to show lower levels of trust. Moreover, it seems that the quality of democracy may negatively influence the levels of trust in the media too. \cite{muller2013mechanisms}, for instance, found that there is a negative correlation between democratic quality, measured by the Freedom House Index, and trust in the media, \cite{tsfati2014individual} showed that this negative correlation becomes insignificant when other factors are taken into account. Similarly, \cite{fawzi2021concepts} contended that ``it is not democracy per se that influences levels of trust, but rather the changes in values and economic prosperity that come along with it.''.

It is also worth mentioning that trust in the media must not be considered a value in itself. People should trust the media if and when there are solid reasons to do so. But assessing the media's trustworthiness, as we will point out further on, is a quite complicated task, as it involves analyzing a variety of different factors. Following Norris, we contend that our political communities require citizens capable of critically examining if their judgments about trustworthiness are based on sound or erroneous predictors, that is, if there are good reasons to trust or to distrust \cite{norris2022praise}. The point here is to avoid what Norris calls sceptical mistrust, cynical mistrust, and credulous trust. People should not mistrust the media without a prior assessment of their work, nor trust them regardless of what they do. To assess if institutions are worthy of trust, Norris suggests that decisions to trust should be internally consistent, meaning that they should be based upon deliberative logical thinking, and externally consistent, that is, they should be based on an examination of reliable and well-grounded sources of information. The relationship between people and the news got further complicated as some of the contents to which people are exposed are disclosed by a virtually endless amount of sources no longer limited to institutionalized news media outlets. This study aims to examine some of the factors that may help people assess if the content that they consume is reliable or not, in order to identify what are the main factors that explain people’s decision to trust or distrust information on content widely discussed within the public sphere (COVID-19, climate change and gender issues).

\section{Literature Review}
\label{sec:literaturereview}

One of the few things that we know about trust/distrust in the news is that it is both a very relevant and, at the same time, a very complex topic. In this section, we will try to identify what are the factors that are generally identified as predictors of trust and challenge the difficulty associated with determining which are the most relevant ones. Our goal is to outline some research difficulties discussed in the literature on trust and explain the rationale that supports the methodological approach adopted in this study. 

According to the widely cited definition of trust proposed by \cite{mayer1995integrative}, trust ``is the willingness of a party to be vulnerable to the actions of another party based on the expectation that the other will perform a particular action important to the trustor, irrespective of the ability to monitor or control that other party''. The authors argue that this expectation is motivated by two intertwined types of factors: trustworthiness and propensity. The first one is the result of a rational assessment of three different features: 1) the ability of the trustee to perform its task; 2) the benevolence of the trustee and its concern for the well-being of the trustor; and 3) the integrity of the trustee, that is, its adherence to a set of values and its word-deeds consistency. The second one (propensity) results from a more emotional process that refers to a general habit or disposition that is formed over time as a consequence of past experiences that people go through in their socialization process. It particularly involves the opinions of relevant people in one’s life such as relatives, teachers, friends; level of education, and other hard to count cultural factors \cite{baer2018not,liu2023factors,mcallister1995affect,vanknippenberg2018reconsidering}.

Most academic research on trust has focused on trustworthiness and has tried to identify and measure the relevance of the various antecedents of trust \cite{isaeva2015why}. According to the authors, this trend reflects, on the one hand, the enormous influence the positivist epistemological paradigm has had over the research on trust and, on the other hand, the difficulty of navigating through the uncountable factors that shape people’s dispositions to trust, many of which people are not aware of. On top of all these research difficulties it is worth mentioning that these more emotional factors influencing trust end up impacting and moderating people's rational assessments of the trustee's ability, benevolence, and integrity \cite{baer2018not}. Among others, these could be some of the reasons why it has been difficult to provide a general explanation of the phenomenon.
Recent research found that ``it is increasingly unclear what people are thinking about when asked about their trust in news media (types) in general'' \cite{tsfati2022going}. Having examined how audiences understand trust in the media, \cite{knudsen2022how} concluded that people associate it with four main ideas: 1) bias; 2) trustfulness; 3) thoroughness and professionalism; and 4) independence and objectivity. And \cite{kohring2007trust} suggested that by trusting the news media, people would trust in the selection of content, specific approach, sources and frame. People would trust news pieces because of 1) the nature of the message itself, 2) the source of the message, or 3) the channel or medium used to disseminate the message.

\cite{stromback2020news} recommend that future research should specify the objects of trust, in order to help the respondents of surveys, focus groups or experiments to identify clearly what they are being asked. Moreover, instead of focusing on trust in the media, researchers should focus on ``trust in the information coming from news media''. Nonetheless, the authors warn us that research also shows that ``news media users do not have at their disposal the resources and capabilities to evaluate thoroughly the reliability of news'' and that their assessment is often influenced by other factors not necessarily associated with the perception they have formed of how trustworthy media performance is (\cite{stromback2020news} p. 141). Studies focused on motivated reasoning, hostile media and confirmation bias are particularly illustrative of the role played by these other factors (predispositions) on trust. People tend to perceive conflicting information as wrong and to classify information that fits their ideology and beliefs as truthful and reliable, regardless of their truth value \cite{pereira2023identity,savolainen2022what,thaler2024fake, vegetti2020impact}. 

The more politically engaged individuals are, the more likely it is for them to be affected by this bias \cite{gunther2017who,nyhan2010when}. Bearing this in mind, it is important not to neglect the fact that media trust may vary depending on the topics covered. Both \cite{stromback2020news} and \cite{tsfati2022going} believe that the little attention paid to these two features of trust in news media helps explain why regardless of the extensive scientific research, ``our knowledge of news media trust [is] more limited that appears at first glance'' (\cite{stromback2020news}, p. 145).

\cite{gunther2017who} also found out that people tend to put forward a defensive mechanism when exposed to content sourced by hostile media brands. In these occasions, positive variations on the content tend to have little effect on their assessments of trust, whereas negative variations tend to reinforce their distrust. And the same happens when content is sourced by friendly media. In this case, trust in congenial content is reinforced, whereas hostile content ends up producing little effect on trust. However, these results cannot be generalized to all sorts of profiles. People’s responses to this hostile media effect also vary depending on the group’s ``relative status in society'': people holding well-established opinions tend to be less affected by the source and content compared to those holding less popular views. In a nutshell, perceptions of media bias are strongly correlated with the sensation people have that their social group's status is under threat. This is what authors like \cite{inglehart2016trump} call cultural backlash. 

Furthermore, \cite{molina2024does} found that individuals who are more inclined to distrust their fellows are more likely to place trust in content generated or moderated by machines. Humans are likely to be influenced by a series of factors like political ideology or religious beliefs that would compromise their objectivity and fairness. This was particularly the case with conservatives, who, according to \cite{samples2019why}, were more likely to believe that big tech companies like Facebook or Twitter unfairly classify and censor right conservative content. Congruent with \cite{gunther2017who}, these findings raise the question of how people would react if they were not told if the content they are presented to was produced by human beings or by chatbots.

As people have been increasingly exposed to misinformation campaigns on social media (authors), it makes sense to try to identify the elements within the messages themselves that impact their levels of trust. The purpose is to see if variations introduced on the headlines and body of the text (stances, sources, clickbait and toxic language) affects people’s trust in content related to controversial or divisive topics (COVID-19 vaccines, climate change and gender issues). By shifting the focus placed on news media sources, as in \cite{gunther2017who}'s experiment, to the content of messages themselves, this study aims to evaluate readers' ability to identify fake stories based solely on the content, avoiding the hostile media effect caused by authorship. In order to do this, an online experiment was conducted where participants (800 in Spain and 800 in the UK) were asked to read three fake items and rate their level of trust on a scale from 1 (not true) to 8 (true). This proceeding also prevented participants from not knowing what exactly was the object of trust they were asked to evaluate.  
The pieces used a combination of factors, including stance (favourable, neutral, or against the narrative), presence of toxic expressions, clickbait titles, and sources of information to test which elements influenced people's responses the most. Half of the pieces were produced by humans and the other half by ChatGPT. Through this experiment, the study aims to answer the following research questions: 

\begin{itemize}
    \item \textbf{RQ1}: What elements/factors within the message itself (topic covered, stance regarding the topic, presence of sources of information, inclusion of clickbaits in headlines and use of toxic language) influence people’s levels of trust the most?
    \item \textbf{RQ2}: Is Generative AI (ChatGPT) capable of writing disinformation pieces at the same deception level as humans? 
    \item \textbf{RQ3}: Is generative AI performing similarly in different languages and to different audiences (Spain and UK)? 

\end{itemize}

\section{Methodology}
\label{sec:methodology}

The experiment was designed to identify factors that affect people’s trust level in news articles. In addition to that, during the experiment design, we decided to include the same factor combinations into news stories written both by humans and by ChatGPT \cite{ouyang2022training}. In this way, we could identify the extent to which there would be any differences in trust factors depending on authorship. Experiment results would then be used to create an automated tool aimed at real-world flagging of untrustworthy content in online media.

The core objective of our human evaluation experiment was to see which factors increased the chances of people trusting fake stories. To do this, we chose the three most popular current disinformation narratives circulating at the time on social media and news media. Popularity was measured by capturing fact checks provided by one of the most credited Spanish fact-checkers (Newtral), which were first encoded as numerical representations, also known as embeddings in the natural language processing literature \cite{mikolov2013distributed} and then clustered, or grouped together according to some distance metric in euclidean space. We used the DBScan algorithm \cite{ester1996density}. Then, these clusters were ranked by size, and sufficiently representative and diverse clusters were manually selected, and then passed to ChatGPT for summarization. Note that the goal was not to determine which were the factors that made citizens trust news, but the factors that made them trust fake stories. Otherwise, we would have also created real stories to compare them. An example of a cluster of stories (in their original language of publication, Spanish) covering the same topic is provided below, alongside the identified shared narrative with ChatGPT:
\begin{quote}
\begin{itemize}
    \item Story 1: En el manicomio donde nos quieren enjaular a la fuerza, esta es la nueva definici\'on de mujer: ``Una adulta que vive y se identifica como mujer, aunque haya nacido con un sexo diferente''.
    
    \begin{small}\texttt{[Translation: In the asylum where they want to force us into cages, this is the new definition of woman: ``An adult who lives and identifies as a woman, even if she was born with a different sex''.]}
    \end{small}
    
    \item Story 2: \textquestiondown{}\textit{El diccionario de Cambridge cambi\'o la definici\'on de ``mujer'' a ``adulto que se identifica'' como tal ``incluso si naci\'o de un sexo diferente''?}
    
    \begin{small}\texttt{[Translation: Did the Cambridge Dictionary change the definition of ``woman'' to ``adult who identifies'' as such ``even if born of a different sex''?]}
    \end{small}
    
    \item Story 3: \textit{Cambridge Dictionary cambi\'o las definiciones de ``mujer'' y ``hombre''. Aquellos que son hombres pero creen que son mujeres ahora podr\'an usar el ba\~no de mujeres debido a la ley de Maranh\~ao}
    
    \begin{small}\texttt{[Translation: Cambridge Dictionary changed the definitions of ``woman'' and ``man''. Those who are men but believe they are women will now be able to use the women's bathroom due to the Maranhão law]}
    \end{small}
    
    \item Summarized narrative: La narrativa compartida entre los bulos es que hay una supuesta redefinici\'on de la definici\'on de ``mujer'' que incluye a personas que se identifican como tal, independientemente de su sexo biol\'ogico.
    
    \begin{small}\texttt{[Translation: The narrative shared among the hoaxes is that there is a supposed redefinition of the definition of ``woman'' that includes people who identify as such, regardless of their biological sex.]}
    \end{small}
\end{itemize}
\end{quote}

\subsection{Experimental Design}
\label{sec:expdesign}

Each participant, after completing sections on sociodemographic questions and the impact of misinformation, received three ``cards''. A card consisted of a text with two paragraphs and a title (longer versions were shortened for the online survey format), presented in a neutral context that does not resemble a media outlet. Each card was constructed using a specific combination of factors (detailed below). The design focused on having participants evaluate textual factors. After reading each text, participants were asked to rate the credibility of the content on a scale from 1 (completely disbelieve) to 7 (completely believe).

Each participant was presented with one text from each of the following 3 narratives. With 800 surveys per country, this means we obtained evaluations for 2,400 cards in Spain and 2,400 in the United Kingdom, for a total of 4,800 data points. Below we list the description of each narrative as they appeared in the participants' data.

\begin{itemize}
    \item N1 - \textbf{Electric Cars}: Electric cars emit harmful electromagnetic radiation. 
    \item N2 - \textbf{Gender}: Cambridge Dictionary modified its definitions of woman and man, leading the British government to adapt its citizen registry based on the new definition. 
    \item \textbf{N3 - Vaccines}: The COVID-19 vaccine created a new blood type.

\end{itemize}

The factors considered in constructing the cards for each narrative were:
\begin{itemize}
    \item Headline and text stance towards the narrative (supportive/opposing/neutral, with headline and text maintaining consistent positioning). 
    \begin{itemize}
        \item Neutral: Devoid of emotion and bias toward the narrative. Text lacks adjectives, resembling wire service news.
        \item Opposing (contrary) narrative: Takes a clear stance against the narrative with partial text. Uses adjectives and appeals to emotion.
        \item Supporting (favour) narrative: Takes a clear stance supporting the narrative with partial text. Uses adjectives and appeals to emotion.
    \end{itemize}
    \item Clickbait presence/absence in headline: For cards either supporting or opposing the narrative, a clickbait headline variant is included.
    \item Toxicity presence/absence: For cards either supporting or opposing the narrative, a toxic phrase variant is included, containing attacks or discrimination against specific groups.
    \item Data/expert citations presence/absence: For neutral cards, a variant includes relevant data about the narrative.
\end{itemize}

These factors required creating 8 cards per narrative in each country to cover all possible combinations (Supporting, Opposing, Neutral, Supporting-toxic, Opposing-toxic, Neutral-data, Supporting-clickbait, Opposing-clickbait). Because of this, there were 24 cards per country. Additionally, for each narrative, cards were created by both human authors and AI (similarly to the rest of the experiment, using ChatGPT). This doubles the total to 16 cards per narrative (8 human-authored, 8 AI-authored), resulting in 48 cards per country. As cards are randomly distributed in the online survey and each person responds to three cards, each card would receive responses from 50 participants per country. In terms of the participants’ demographic information, a total of 1,601 participants took part in the survey, with ages that ranged between 18 and 74 (mean 45).

\section{Findings}
\label{sec:findings}

We initially investigated whether the level of trust in news articles was influenced by factors related to news articles such as the writers, narratives, headlines, toxic phrases, clickbait titles, and quotes from experts. Then, we examined whether information related to participants’ background such as their nationality, age, gender, education level, political ideology, trust in news, and news consumption would influence their level of trust in news articles.

\subsection{The effect of narratives, headlines, and writers}
\label{sec:narheadwriters}

\begin{figure}[!h]
    \centering
    \includegraphics[width=0.9\textwidth]{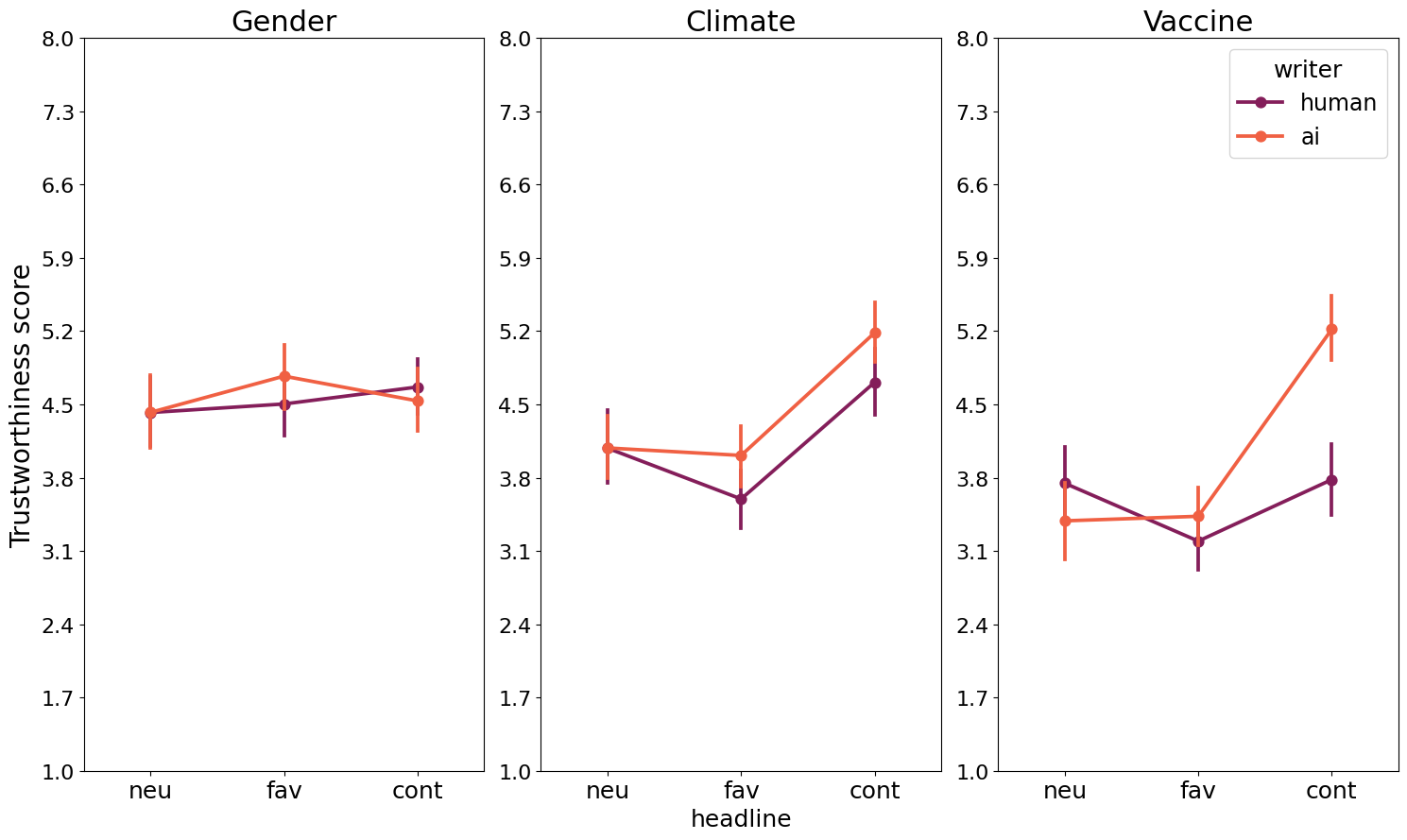}
    \caption{Trustworthiness scores by narratives (gender, climate, and vaccine), headlines (neutral, favour, and contrary), and writers (AI and Human).}
    \label{fig:narheadwriters}
\end{figure}

Figure \ref{fig:narheadwriters} shows the trustworthiness scores of news articles presented in surveys according to the narratives (gender, climate, and vaccine), headlines (neutral, favour, and contrary), and writers (AI and human).  We excluded responses (i.e., ``I don’t know'' and ``I prefer not to answer'') that did not indicate how much participants trusted the articles. In the rest of the analyses in this section, we analysed the trustworthiness scores using a mixed-effects regression model implemented in R \cite{rcoreteam2020r} using the \texttt{lme4} package \cite{bates2015fitting}. The significance of the effects was examined by a likelihood ratio test using the \texttt{anova()} function in R \cite{jaeger2008categorical}. In subset models, the effect of interest was removed at a time, starting with an interactive result. We compared the significance of an effect by comparing the full model and a subset model without the effect of interest. If the absence of the effect significantly affected the model fit, it remained in the model. The maximum random effects structure that converged was included in all models. The model examining the influence of narratives, headlines, and writers included fixed effects for Narratives (gender, climate, or vaccine), Headlines (neutral, favour, or contrary), Writers (AI or human) and their interaction effects. The model included a random intercept of Participants. The result of a likelihood ratio test shows that the interaction effects of Narratives by Headlines by Writers significantly affected the model fit $(\chi^2(4) = 25.22, p < .01)$; hence, they remained in the model.
 
Table \ref{tab:narheadwriters} shows the summary of the fixed effects in the final model predicting the trustworthiness scores according to narratives, headlines, and writers. Regarding the intercept of the model, articles about climate that have a neutral stance and were written by humans were positive and significant, indicating that participants showed a relatively higher baseline trust level on the articles. The simple fixed effect of Headlines (favour) was negative and Headlines (contrary) was positive, and they were both significant. This indicates that the trustworthiness scores were significantly lower when articles had a favourable stance (e.g., ``Electric cars emit harmful electromagnetic radiation'') and higher when they had a contrary stance towards fake news (e.g., ``Study dismisses electric cars as harmful to health'') compared to when articles had a neutral stance (e.g., ``EU investigates whether electric cars emit harmful electromagnetic radiation''). 

Regarding gender topics, the interaction effects of Narratives (gender) by Headlines (favour) were positive and significant, indicating that trustworthiness scores were significantly higher when articles about gender had a favourable stance (e.g., ``The British Administration embraces transgender ideology'') than when they had a neutral stance (e.g., ``Change in the definition of \textit{Man} and \textit{Woman} will require citizens to declare their self-identified gender''). In other words, although participants generally placed less trust in fake news articles with a favourable stance than those with a neutral stance, they trusted those with a favourable stance as much as those with a neutral stance when the articles addressed gender-related topics. 

When it comes to pieces focused onCOVID-199, the interaction effects of Narratives (vaccine) and Headline (contrary) were negative and significant, indicating the trustworthiness scores of articles on vaccines were significantly lower when they had a contrary stance (e.g., ``The COVID-19 vaccine does not alter blood type'') than when they had a neutral stance (e.g., ``Scientific discussion on the COVID-19 vaccine and a new blood group''). Overall, participants generally trusted fake news articles with a contrary stance more than those with a neutral stance. However, they trusted vaccine-related articles less even when the articles had a contrary stance. Lastly, the interaction effects of Narratives (vaccine) by Headlines (contrary) by Writers (AI) were positive and significant, suggesting that, for the articles on vaccines that had a contrary stance, people trusted the articles significantly more when they were written by AI than when they were written by humans.

\begin{table}[htbp]
  \centering
  \caption{Summary of the fixed effects in the final model predicting the trustworthiness scores according to Narratives, Headlines, and Writers.}
  \label{tab:trustworthiness}
  \begin{tabular}{lcccc}
    \toprule
    \textbf{Fixed Effect} & \textbf{Estimate} & \textbf{Standard Error} & \textbf{t-value} & \textbf{P-value} \\
    \midrule
    Intercept & \textbf{4.11} & \textbf{0.16} & \textbf{25.12} & \textbf{$<.001$***} \\
    Gender & 0.24 & 0.22 & 1.12 & .26 \\
    Vaccine & -0.36 & 0.22 & -1.65 & .10 \\
    \textbf{Favour} & \textbf{-0.56} & \textbf{0.2} & \textbf{-2.77} & \textbf{$<.01$**} \\
    \textbf{Contrary} & \textbf{0.66} & \textbf{0.21} & \textbf{3.21} & \textbf{$<.01$**} \\
    AI & 0.04 & 0.21 & 0.18 & .86 \\
    \textbf{Gender \& Favour} & \textbf{0.59} & \textbf{0.28} & \textbf{2.09} & \textbf{.04*} \\
    Vaccine \& Favour & 0.04 & 0.28 & 0.15 & .88 \\
    Gender \& Contrary & -0.5 & 0.28 & -1.76 & .08 \\
    \textbf{Vaccine \& Contrary} & \textbf{-0.62} & \textbf{0.28} & \textbf{-2.16} & \textbf{.03*} \\
    Gender \& AI & 0.03 & 0.3 & 0.1 & .92 \\
    Vaccine \& AI & -0.31 & 0.3 & -1.01 & .31 \\
    Favour \& AI & 0.36 & 0.28 & 1.29 & .20 \\
    Contrary \&AI & 0.24 & 0.28 & 0.85 & .39 \\
    Gender \& Favour \& AI & -0.06 & 0.4 & -0.14 & .89 \\
    Vaccine \& Favour \& AI & 0.09 & 0.39 & 0.22 & .82 \\
    Gender \& Contrary \& AI & -0.16 & 0.39 & -0.42 & .67 \\
    \textbf{Vaccine \& Contrary \& AI} & \textbf{1.36} & \textbf{0.39} & \textbf{3.44} & \textbf{$<.001$***} \\
    \bottomrule
    \multicolumn{5}{p{0.95\linewidth}}{\small The reference group (intercept): Climate, neutral, Human writer, \textbf{R code for the final model}: \texttt{lmer(score $\sim$ narr\_type*headline*ai\_human + (1|Subject), data=data)}.} \\
  \end{tabular}
\end{table}

\subsection{The effect of toxicity, clickbait, and experts’ quotes}
\label{sec:toxicclickbexperts}
 
\begin{figure}[!h]
    \centering
    \includegraphics[width=0.9\textwidth]{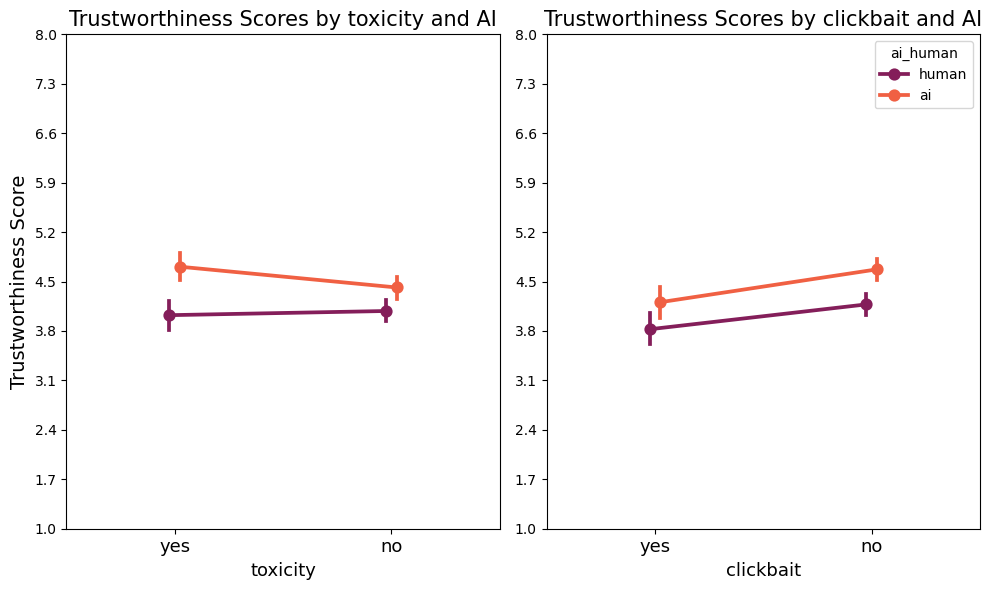}
    \caption{Figure 2. Trustworthiness according to toxicity and writers (left) and clickbait and writers (right).}
    \label{fig:toxicclickbexperts}
\end{figure}

In addition, we investigate how much people trusted articles based on the fact that they contained toxic phrases, clickbait titles, and experts’ quotes. First, we examined the effect of toxicity and clickbait as well as their interaction effects with writers on the trustworthiness score. Figure \ref{fig:toxicclickbexperts} shows the trustworthiness scores according to toxicity and writers (left) and clickbait and writers (right). The model included the fixed effects of Toxicity (Toxic or No Toxic), Clickbait (Clickbait or No Clickbait) and their interaction with the Writers (AI or Human). A likelihood ratio test shows that the interaction effects of Clickbait by the Writers were not significant ($\chi^2(1) = 0.22, p= .64$) and they were removed from the model. The interaction effects of Toxicity by Writers significantly affected the model fit ($\chi^2(1) = 4.68, p= .03$) and they remained in the model.

Table \ref{tab:toxicclickbexperts} shows the fixed effects of the final model predicting the trustworthiness scores according to Toxicity, Clickbait, and Writers. The intercept, representing articles without toxic phrases or clickbait titles and written by humans, was positive and significant. This suggests that people had a relatively higher baseline trust in these articles. The simple effects of Clickbait and Toxicity were negative and significant, showing that people trusted news articles that had clickbait titles or toxic phrases less than the articles that did not. The simple effect of AI was positive and significant, suggesting that articles were perceived as more trustworthy when written by AI compared to those written by humans. The interaction effects between Toxicity (Toxic) and Writers (AI) were positive and significant, indicating that while articles containing toxic phrases were generally trusted less, participants showed higher trust when articles with toxicity were written by AI.

\begin{table}[htbp]
  \centering
  \caption{Summary of the fixed effects in the final model predicting the trustworthiness scores according to Clickbait, Toxicity and Writers.}
  \begin{tabular}{lcccc}
    \toprule
    \textbf{Fixed Effect} & \textbf{Estimate} & \textbf{Standard Error} & \textbf{t-value} & \textbf{P-value} \\
    \midrule
    \textbf{Intercept} & \textbf{4.27} & \textbf{0.09} & \textbf{46.67} & \textbf{$<.001$***} \\
    \textbf{Clickbait} & \textbf{-0.46} & \textbf{0.1} & \textbf{-4.68} & \textbf{$<.001$ ***} \\
    \textbf{Toxicity} & \textbf{-0.3} & \textbf{0.13} & \textbf{-2.33} & \textbf{.02 *} \\
    \textbf{AI} & \textbf{0.35} & \textbf{0.1} & \textbf{3.58} & \textbf{$<.001$ ***} \\
    \textbf{AI} \& \textbf{Toxicity} & \textbf{0.36} & \textbf{0.17} & \textbf{2.17} & \textbf{.03 *} \\
    \bottomrule
    \multicolumn{5}{p{0.95\linewidth}}{\small The reference group (intercept): No clickbait, No Toxicity, Human writer, \textbf{R code for the final model}: \texttt{lmer(score $\sim$ clickbait + toxicity*ai\_human + (1|Subject), data=data)} \textbf{R code for the full model}: \texttt{lmer(score $\sim$ clickbait*ai\_human + toxicity*ai\_human + (1|Subject), data=data)}} \\
  \end{tabular}
  \label{tab:toxicclickbexperts}  
\end{table}

\begin{figure}[!h]
    \centering
    \includegraphics[width=0.9\textwidth]{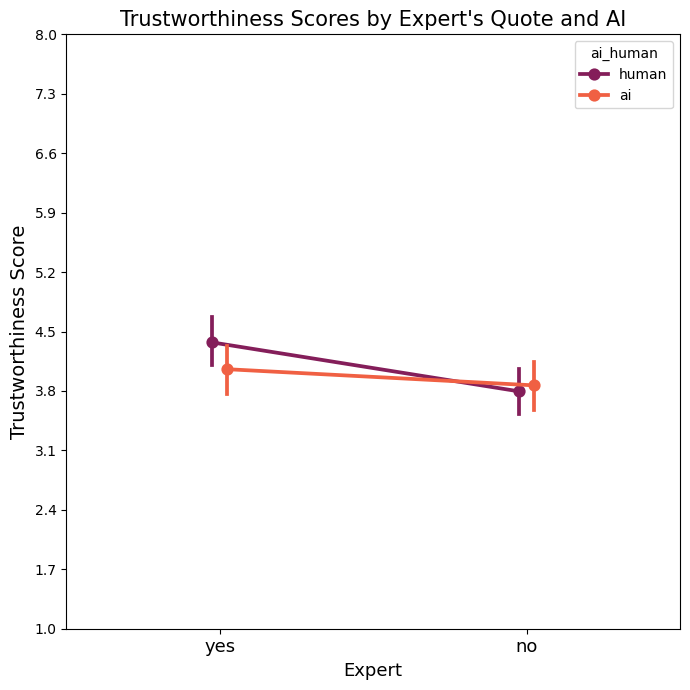}
    \caption{Trustworthiness scores according to experts’ quotes and writers.}
    \label{fig:exptquoteswriters}
\end{figure}

Furthermore, we investigated whether the inclusion of expert quotes in articles had a significant impact on people’s trust in news articles. Figure \ref{fig:exptquoteswriters} shows the trustworthiness scores according to experts’ quotes and writers. The model included the fixed effect of Experts (Quotes or No quotes) and Writers (AI or Human). A likelihood ratio test showed that the interaction effect of Experts by Writers $(\chi^2(1) = 2.46, p = .12)$ and a single effect of Writers $(\chi^2(1) = 1.24, p= .27)$  did not have a significant effect on model fit; hence, they were removed from the model. However, removing the simple effect of Experts significantly affected the model fit $(\chi^2(1) = 3.99, p= .05)$; thus, it remained in the model. Table \ref{tab:exptquoteswriters} shows the fixed effects of the final model predicting the trustworthiness score according to Experts and Writers. The intercept, articles that do not have experts’ quotes and were written by humans, was positive and significant, indicating that people trusted these articles with relatively high trust levels. The single effect of Experts (news that has experts’ quotes) was positive and significant, indicating that having experts’ quotes in articles significantly increased the trustworthiness score.
 
\begin{table}[htbp]
  \centering
  \caption{Summary of the fixed effects in the final model predicting the trustworthiness scores according to Experts.}
  \label{tab:experts}
  \begin{tabular}{lcccc}
    \toprule
    \textbf{Fixed Effect} & \textbf{Estimate} & \textbf{Standard Error} & \textbf{t-value} & \textbf{P-value} \\
    \midrule
    \textbf{Intercept} & \textbf{3.91} & \textbf{0.11} & \textbf{37.13} & \textbf{$<.001$ ***} \\
    \textbf{Expert} & \textbf{0.26} & \textbf{0.13} & \textbf{2.0} & \textbf{.05 *} \\
    \bottomrule
    \multicolumn{5}{p{0.95\linewidth}}{\small The reference group (intercept): No expert, Human writer, \textbf{R code for the final model}: \texttt{lmer(score $\sim$ expert + (1|Subject), data=data)} \textbf{R code for the full model}: \texttt{lmer(score $\sim$ expert*ai\_human + (1|Subject), data=data)}} \\
  \end{tabular}
  \label{tab:exptquoteswriters}
\end{table}

\subsection{Spain and the UK}
\label{sec:spainuk}
 
\begin{figure}[!h]
    \centering
    \includegraphics[width=0.9\textwidth]{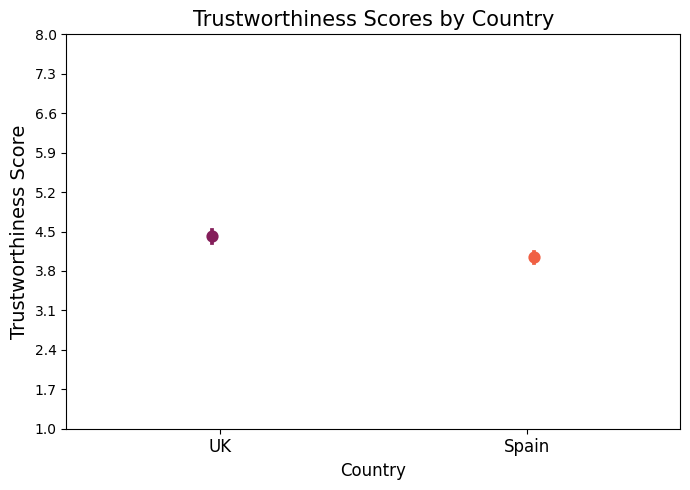}
    \caption{Trustworthiness scores according to participants' country.}
    \label{fig:countrytrustscores}
\end{figure}

We also checked if there were differences between how many participants from Spain and the UK trusted articles in the surveys. Figure \ref{fig:countrytrustscores} shows the trustworthiness scores of participants from Spanish and the UK. The model predicting the trustworthiness between the two participant groups included the fixed effect of Countries (Spain or the UK). Removing the fixed effect of Countries significantly affected the model fit ($\chi^2(1) = 12.98, p< .01$); hence, it remained in the model. Table \ref{tab:countrytrustscores} shows the fixed effect in the final model predicting the trustworthiness scores between participants from Spain and the UK. The intercept, of participants from the UK, was positive and significant, suggesting that British participants had a relatively higher baseline trust in the articles. The single effect of Countries (Spain) was negative and significant, showing that Spanish participants trusted the articles significantly less compared to British participants.

\begin{table}[htbp]
  \centering
  \caption{Summary of the fixed effects in the final model predicting the trustworthiness scores according to country.}
  \label{tab:countrytrustscores}
  \begin{tabular}{lcccc}
    \toprule
    \textbf{Fixed Effect} & \textbf{Estimate} & \textbf{Standard Error} & \textbf{t-value} & \textbf{P-value} \\
    \midrule
    Intercept & \textbf{4.4} & \textbf{0.18} & \textbf{24.24} & \textbf{$<.001$ ***} \\
    \textbf{Spain} & \textbf{-0.37} & \textbf{0.10} & \textbf{-3.61} & \textbf{$<.001$ **} \\
    \bottomrule
    \multicolumn{5}{p{0.95\linewidth}}{\small The reference group (intercept): UK, \textbf{R code for the final model}: \texttt{lmer(score $\sim$ + (1|Subject) + (1|narr\_type) , data=data)} \textbf{R code for the full model}: \texttt{lmer(score $\sim$ country + (1|Subject) + (1|narr\_type), data=data)}.} \\
  \end{tabular}
\end{table}

\subsection{Gender}
\label{sec:gender}

\begin{figure}[!h]
    \centering
    \includegraphics[width=0.9\textwidth]{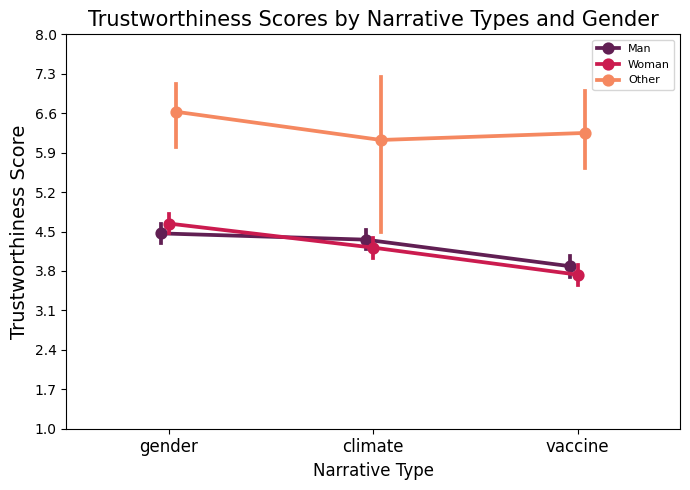}
    \caption{Trustworthiness scores according to participants' gender (man, woman, and others) and narratives (gender, climate, and vaccine).}
    \label{fig:gendernarratives}
\end{figure}
 
When it comes to gender, the study also looked at the differences between how much participants with different genders trusted articles in the surveys. Figure \ref{fig:gendernarratives} shows the level of trustworthiness score by Gender (man, woman, and others) and Narratives (gender, climate, and vaccine). The model included fixed effects for Narratives (gender, climate, or vaccine), Gender (man, woman, or others), and their interaction effects. A likelihood ratio test showed that removing the interactive effect of Narratives by Gender significantly affected the model fit ($\chi^2(3) = 8.60, p = .04$); hence, they remained in the model. The model included random intercepts of Narratives and Participants.

Table \ref{tab:gendernarratives} shows the summary of the fixed effects in the final model predicting the trustworthiness scores according to narratives and gender. The intercept of the model, articles about climate and women, was positive and significant. This shows that women had a relatively higher baseline trust in the articles. A simple effect of Gender (other) was positive and significant, indicating that individuals identifying as "other" trusted news articles more than women (average of the aggregate trustworthiness scores: women: 4.28, men: 4.24, other: 6.32). Furthermore, the interaction effects of Gender (man) and Narratives (gender) were negative and significant, indicating that men particularly trusted articles about gender less than other articles. 

\begin{table}[htbp]
  \centering
  \caption{Summary of the fixed effects in the final model predicting the trustworthiness scores according to Gender and Narratives.}
  \label{tab:gendernarratives}
  \begin{tabular}{lcccc}
    \toprule
    \textbf{Fixed Effect} & \textbf{Estimate} & \textbf{Standard Error} & \textbf{t-value} & \textbf{P-value} \\
    \midrule
    \textbf{Intercept} & \textbf{4.18} & \textbf{0.09} & \textbf{44.18} & \textbf{$<.001$ ***} \\
    Man & 0.17 & 0.13 & 1.31 & .19 \\
    \textbf{Other} & \textbf{1.95} & \textbf{0.84} & \textbf{2.33} & \textbf{.02*} \\
    \textbf{Gender} & \textbf{0.43} & \textbf{0.1} & \textbf{4.25} & \textbf{$<.001$ ***} \\
    \textbf{Vaccine} & \textbf{-0.44} & \textbf{0.1} & \textbf{-4.37} & \textbf{$<.001$ ***} \\
    \textbf{Man \& Gender} & \textbf{-0.36} & \textbf{0.14} & \textbf{-2.57} & \textbf{.01*} \\
    Other \& Gender & 0.07 & 0.88 & 0.07 & .94 \\
    Man \& Vaccine & -0.03 & 0.14 & -0.18 & .86 \\
    Other \& Vaccine & 0.57 & 0.88 & 0.64 & .52 \\
    \bottomrule
    \multicolumn{5}{p{0.95\linewidth}}{\small The reference group (intercept): Climate, women, \textbf{R code for the final model/R code for the full model}: \texttt{lmer(score $\sim$ gender*narr\_type + (1|Subject) + (1|narr\_type), data=data)}.} \\
  \end{tabular}
\end{table}

\subsection{Age}
\label{sec:age}

\begin{figure}[!h]
    \centering
    \includegraphics[width=0.9\textwidth]{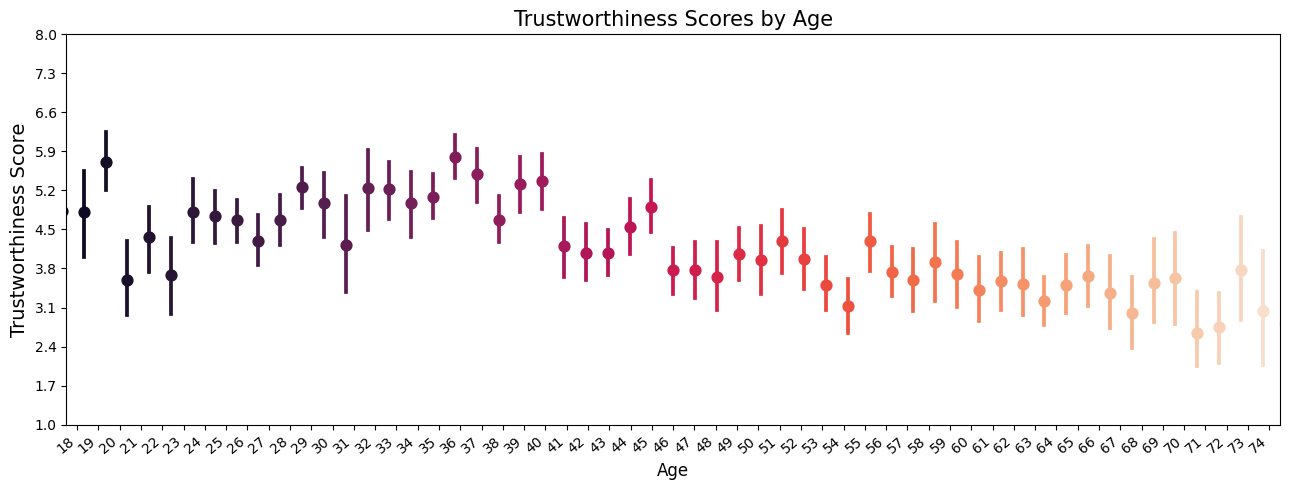}
    \caption{Trustworthiness scores according to participants' age.}
    \label{fig:age}
\end{figure}

We examined whether the trustworthiness score was significantly affected by participants' age. Figure \ref{fig:age} shows the level of trustworthiness score by age. The model included the fixed effect for age. A likelihood ratio test showed that removing the simple effect of Age significantly affected the model fit ($\chi^2(1) = 125.94, p< .001$); hence, it remained in the model. The model included random intercepts of Narratives and Participants. Table \ref{tab:age} presents a summary of the fixed effects in the final model predicting trustworthiness scores based on participants' age. The simple effect of Age was negative and significant, suggesting that as people get older, their trust in fake news articles decreases.

\begin{table}[htbp]
  \centering
  \caption{Summary of the fixed effects in the final model predicting the trustworthiness scores according to Age.}
  \label{tab:age}
  \begin{tabular}{lcccc}
    \toprule
    \textbf{Fixed Effect} & \textbf{Estimate} & \textbf{Standard Error} & \textbf{t-value} & \textbf{P-value} \\
    \midrule
    \textbf{Intercept} & \textbf{5.97} & \textbf{0.23} & \textbf{25.85} & \textbf{$<.001$ ***} \\
    \textbf{Age} & \textbf{-0.04} & \textbf{0.00} & \textbf{-11.50} & \textbf{$<.001$ ***} \\
    \bottomrule
    \multicolumn{5}{p{0.95\linewidth}}{\small The reference group (intercept): Climate, women, \textbf{R code for the final model/R code for the full model}: \texttt{lmer(score $\sim$ age + (1|Subject) + (1|narr\_type), data=data)}} \\
  \end{tabular}
\end{table}

\subsection{Education Level}
\label{sec:education}

\begin{figure}[!h]
    \centering
    \includegraphics[width=1\textwidth]{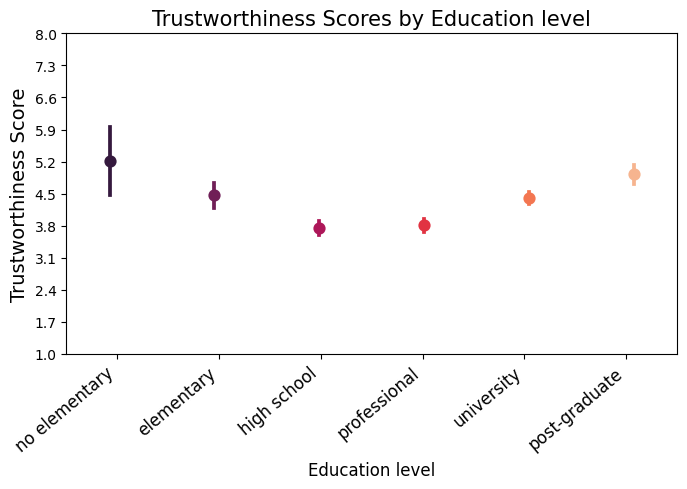}
    \caption{Trustworthiness score according to participants' education level}
    \label{fig:education}
\end{figure}

We also tested whether participants' education level would significantly affect the trustworthiness score. Figure \ref{fig:education} shows the trustworthiness score by participants' education level. The model included the fixed effect of participants' education level. A likelihood ratio test showed that removing the fixed effect significantly affected the model fit ($\chi^2(5) = 66.97, p< .001$); hence, it remained in the model. The model included the random intercepts of Narratives and Participants. Table \ref{tab:education} shows the summary of the final model predicting the trustworthiness scores based on participants' education levels. The intercept, no elementary level, was positive and significant, indicating that participants without an elementary education level scored the news articles relatively high. The fixed effects of high school and professional were negative and significant, indicating that participants with high school and professional level education trusted news articles significantly less than those without elementary level. No other effects were significant.

\begin{table}[htbp]
  \centering
  \caption{Summary of the fixed effects in the final model predicting the trustworthiness scores according to education level.}
  \label{tab:education}
  \begin{tabular}{lcccc}
    \toprule
    \textbf{Fixed Effect} & \textbf{Estimate} & \textbf{Standard Error} & \textbf{t-value} & \textbf{P-value} \\
    \midrule
    \textbf{Intercept} & \textbf{5.1} & \textbf{0.53} & \textbf{9.62} & \textbf{$<.001$ ***} \\
    Elementary & -0.65 & 0.55 & -1.19 & .23 \\
    \textbf{High school} & \textbf{-1.35} & \textbf{0.51} & \textbf{-2.62} & \textbf{$<.01$ **} \\
    \textbf{Professional} & \textbf{-1.31} & \textbf{0.52} & \textbf{-2.54} & \textbf{.01 *} \\
    University & -0.71 & 0.51 & -1.39 & .16 \\
    Post-graduate & -0.21 & 0.52 & -0.39 & .69 \\
    \bottomrule
    \multicolumn{5}{p{0.95\linewidth}}{\small The reference group (intercept): No elementary, \textbf{R code for the final model/R code for the full model}: \texttt{lmer(score $\sim$ Education\_level + (1|Subject) + (1|narr\_type), data=data)}} \\
  \end{tabular}
\end{table}

\subsection{Political Ideology}
\label{sec:political}

\begin{figure}[!h]
    \centering
    \includegraphics[width=0.9\textwidth]{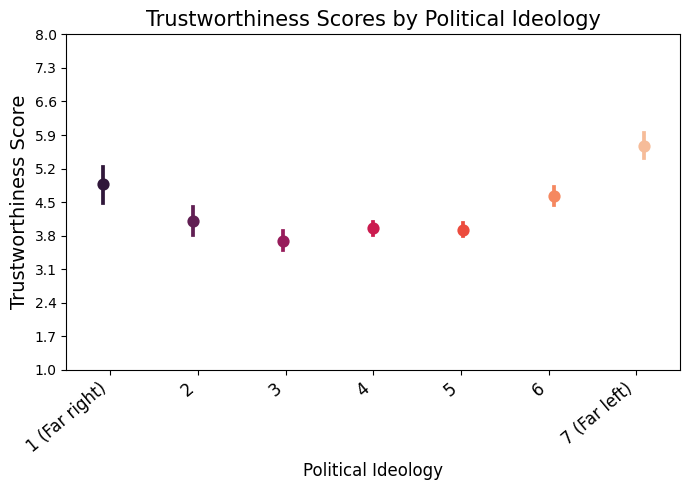}
    \caption{Trustworthiness score according to political ideology}
    \label{fig:political}
\end{figure}

The experiment also examined the extent to which trust was affected by participants' political ideology. Figure \ref{fig:political} shows the trustworthiness score according to their political ideology. The model predicting the trustworthiness scores according to political ideology included the fixed effect of political ideology and random intercepts of Narratives and Participants. A likelihood ratio test showed that removing the fixed effect significantly affected the model fit ($\chi^2(6) = 112.83, p< .001$); hence, it remained in the model. Table \ref{tab:political} shows the final model predicting the trustworthiness scores according to the political ideology. The intercept, centre, was positive and significant, indicating that people with central political views had a relatively higher baseline trust in the articles. The fixed effects of Far-right, Left, and Far-left were positive and significant, indicating that participants with these ideologies trusted the news articles significantly more than people with central political ideology.

\begin{table}[htbp]
  \centering
  \caption{Summary of the fixed effects in the final model predicting the trustworthiness scores according to political ideology.}
  \label{tab:political}
  \begin{tabular}{lcccc}
    \toprule
    \textbf{Fixed Effect} & \textbf{Estimate} & \textbf{Standard Error} & \textbf{t-value} & \textbf{P-value} \\
    \midrule
    \textbf{Intercept} & \textbf{3.94} & \textbf{0.19} & \textbf{20.49} & \textbf{$<.001$ ***} \\
    \textbf{Far-right} & \textbf{0.92} & \textbf{0.27} & \textbf{3.47} & \textbf{$<.001$ ***} \\
    Right & 0.18 & 0.22 & 0.84 & .4 \\
    Moderately right & -0.24 & 0.17 & -1.41 & .16 \\
    Moderately left & -0.01 & 0.14 & -0.09 & .93 \\
    \textbf{Left} & \textbf{0.66} & \textbf{0.16} & \textbf{4.28} & \textbf{$<.001$ ***} \\
    \textbf{Far-left} & \textbf{1.69} & \textbf{0.19} & \textbf{8.67} & \textbf{$<.001$ ***} \\
    \bottomrule
    \multicolumn{5}{p{0.95\linewidth}}{\small The reference group (intercept): Centre, \textbf{R code for the final model/R code for the full model}: \texttt{lmer(score $\sim$ Poligical\_ideology + (1|Subject) + (1|narr\_type), data=data)}} \\
  \end{tabular}
\end{table}

\subsection{Trust in Media}
\label{sec:mediattrust}

\begin{figure}[!h]
    \centering
    \includegraphics[width=0.9\textwidth]{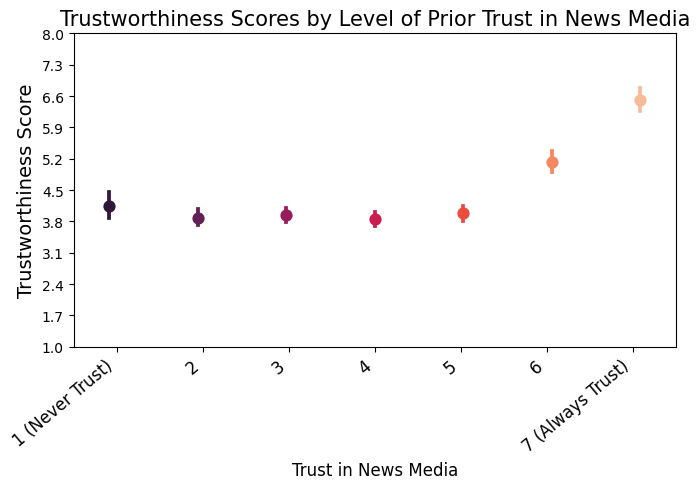}
    \caption{Trustworthiness scores according to participants' level of prior trust in news media}
    \label{fig:mediatrust}
\end{figure}

We also checked if participants' level of trust in news media affected the trustworthiness scores. Figure \ref{fig:mediatrust} shows the trustworthiness scores according to participants' prior trust in news media. The model predicting the scores according to their trust in news media had the fixed effect of trust, which remained in the model ($\chi^2(1) = 79.43, p< .001$). The model also included random intercepts for narrative types and subjects. The fixed effect of trust was positive and significant, indicating that greater trust in news media was associated with higher trust in the news articles.

\begin{table}[htbp]
  \centering
  \caption{Summary of the fixed effects in the final model predicting the trustworthiness scores according to trust in news media.}
  \label{tab:mediatrust}
  \begin{tabular}{lcccc}
    \toprule
    \textbf{Fixed Effect} & \textbf{Estimate} & \textbf{Standard Error} & \textbf{t-value} & \textbf{P-value} \\
    \midrule
    \textbf{Intercept} & \textbf{3.14} & \textbf{0.21} & \textbf{14.99} & \textbf{$<.001$ ***} \\
    \textbf{Trust} & \textbf{0.28} & \textbf{0.03} & \textbf{9.06} & \textbf{$<.001$ ***} \\
    \bottomrule
    \multicolumn{5}{p{0.95\linewidth}}{\small The reference group (intercept): Centre, \textbf{R code for the final model/R code for the full model}: \texttt{lmer(score $\sim$ trust + (1|Subject) + (1|narr\_type), data=data)}} \\
  \end{tabular}
\end{table}

\subsection{News consumption}
\label{sec:newsconsumption}

\begin{figure}[!h]
    \centering
    \includegraphics[width=0.9\textwidth]{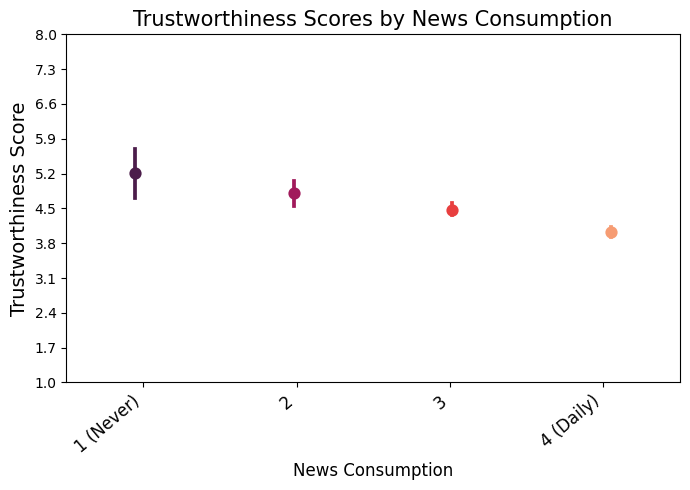}
    \caption{Trustworthiness scores according to News Consumption.}
    \label{fig:consumption}
\end{figure}

Lastly, we examine whether participants' news consumption affected the trustworthiness scores. Figure \ref{fig:consumption} shows trustworthiness scores according to the level of news consumption. The model predicting the trustworthiness scores according to news consumption included the fixed effect of news consumption ($\chi^2(1) = 33.16, p< .001$) and random intercepts of narrative types and subjects. The fixed effect of news consumption was negative and significant, indicating that the more participants consumed news, the less they trusted news articles. 

\begin{table}[htbp]
  \centering
  \caption{Summary of the fixed effects in the final model predicting the trustworthiness scores according to news consumption.}
  \label{tab:consumption}
  \begin{tabular}{lcccc}
    \toprule
    \textbf{Fixed Effect} & \textbf{Estimate} & \textbf{Standard Error} & \textbf{t-value} & \textbf{P-value} \\
    \midrule
    \textbf{Intercept} & \textbf{5.68} & \textbf{0.31} & \textbf{18.45} & \textbf{$<.001$ ***} \\
    \textbf{Trust} & \textbf{-0.42} & \textbf{0.07} & \textbf{-5.79} & \textbf{$<.001$ ***} \\
    \bottomrule
    \multicolumn{5}{p{0.95\linewidth}}{\small The reference group (intercept): Centre, \textbf{R code for the final model/R code for the full model}: \texttt{lmer(score $\sim$ consumption + (1|Subject) + (1|narr\_type), data=data)}} \\
  \end{tabular}
\end{table}

\subsection{Summary and Conclusion}

In summary, the findings indicate that how much people trust fake news articles is affected by their stance. People trusted fake articles more when the articles had a contrary stance towards fake news than when they had a neutral stance. On the contrary, when articles had a favourable stance towards fake news, people trusted them less than those with a neutral stance. Moreover, the findings indicate the trustworthiness of news articles is affected by the combination of stance and narratives. People trusted articles about vaccines less even when they had a contrary stance towards the fake news. However, they trusted articles about vaccines with a contrary stance more when it was written by AI than when it was written by humans. Additionally, people trusted articles about gender with a favourable stance rather than a neutral stance. 

Moreover, people trusted articles less when they contained clickbait titles or toxic phrases. When AI wrote the articles with toxic phrases, people trusted the articles more than human-written toxic articles. They also trusted articles including experts' quotes and related data more than ones that did not. The nationality of participants influenced their trustworthiness scores. Spanish participants generally trusted news articles less than British participants. Additionally, men and women showed similar levels of trust in the articles, while individuals identifying as others demonstrated significantly higher trust compared to both men and women. 

The analysis revealed a significant negative relationship between age and the trustworthiness scores of fake news articles. Older participants were found to have lower trust in fake news articles compared to younger participants. Lastly, participants' educational level, political ideology, trust in news media, and news consumption had significant effects on how much they trusted news articles.

\section{Discussion}
\label{sec:discussion}

\subsection{Generative AI’s ability to generate fake news}
\label{sec:aisabilityfakenews}

The main finding of the current study is that Generative AI can produce fake news across various topics and stances with a similar level of effectiveness as human writers. During the survey, participants were asked to rate how much they trusted fake news articles written by either AI or human authors. The analysis indicates that participants trusted fake news written by AI and humans equally, regardless of the narrative (climate, gender, or vaccine) or stance (neutral, favourable, or contrary). Furthermore, participants exhibited greater trust in AI-generated articles than in those written by humans when the topic was vaccines and the stance was contrary.

\subsection{Narrative effect on trust}

It seems that people in general trusted more fake news about gender than they did about climate change and vaccines. This corroborates previous findings that highlight that the subject of news pieces influences people's trust \cite{thaler2024fake, tsfati2022going}. Although all pieces were fake, people trusted more the climate change stories that included negative stances in their headlines and the body of the text than they did when the stances were neutral. These findings suggest that people may feel reluctant regarding the climate change phenomenon and that the traditional neutral and impartial stance typical of journalistic writing seems not to influence people's attitudes towards the contents. People's attitudes regarding gender stories seem to corroborate this hypothesis. In this particular case, the situation reverses: people trusted more stories with a positive stance regarding gender than the ones with negative or neutral stances. 
Taken together these findings also support the thesis according to which predispositions to trust mediate and influence people's rational assessments \cite{mayer1995integrative}.

\subsection{Toxicity, clickbait, sources}

News articles with toxic expressions and clickbait headlines significantly reduced people's trust in the news pieces. This effect was consistent across different narratives and stances, suggesting that readers utilize these stylistic features as a proxy for assessing content credibility. Interestingly, when comparing human versus AI authorship, participants showed higher trust in AI-generated toxic content compared to human-written toxic content, which aligns with \cite{molina2024does}'s findings that people may perceive machines as less influenced by personal biases.

This relationship between structural elements and trust assessments demonstrates how readers' critical evaluations are somewhat independent of ideological predispositions. While confirmation bias certainly plays a role in trust in the media \cite{pereira2023identity, vegetti2020impact}, our findings indicate that certain journalistic features play a critical role regardless of the ideology. The negative effect of clickbait titles on trust ratings (significant at p<.001) indicates that audiences have developed sensitivity to these attention-grabbing tactics, possibly because they are typically linked with lower-quality sources.

Conversely, the presence of expert quotes significantly increased participants' likelihood to trust articles, regardless of their accuracy. This finding reveals an important vulnerability in news consumers' critical evaluation process—professional formatting can potentially override critical assessment, making fake news more effective when it mimics the appearance of serious journalism. This supports \cite{stromback2020news}'s view that news consumers often lack sufficient mechanisms to thoroughly evaluate the reliability of news information.

\subsection{UK vs Spain}

While our experimental design controlled for content exposure by presenting identical narratives to both populations, the observed differences suggest that broader cultural, political, and media-related factors influence how citizens evaluate news credibility (cf. \cite{zuniga2019trust, tsfati2014individual} for discussion on media trust variations across different national contexts). Investigating the motives behind the UK's stronger baseline trust vs the Spanish participants is out of scope for this paper, e.g., by investigating Spanish audiences' sensitivity to news in the media, perhaps due to significant recent political polarization and media fragmentation.

\subsection{Level of Education}

Our analysis of education's impact on trust in fake news revealed a complex relationship that contradicts some common assumptions. A somewhat expected finding was that high school and professional education levels correlated with significantly lower trust in fake news articles compared to those without elementary education. However, contrary to what might be expected from a linear relationship between education and critical awareness towards the media, university and post-graduate educated participants did not show the lowest trust scores. 

The reduced susceptibility to fake news among participants with high school and professional education might indicate that these groups have developed practical critical thinking skills without necessarily adopting the broader trust in institutions that may characterize higher educational attainment. Furthermore, the relatively higher trust among university-educated participants might be an indication of higher susceptibility to confirmation bias. Alternatively, it may indicate greater trust in institutional sources generally, which could extend to trusting fabricated content that effectively mimics legitimate sources.

\subsection{Political Ideology}

Participants identifying with far-right, left, and far-left positions demonstrated significantly higher trust in fake news articles compared to those with centrist views, which aligns with previous works pointing to a relationship between political extremism and patterns in media consumption \cite{mudde2017populism, norris2022praise}. However, our findings are particularly interesting in that the effect was observed at both ends of the political spectrum, suggesting that polarization and ideological intensity \textit{in general} (without considering political adherence) correlates with susceptibility to fake news. Again, one plausible explanation is susceptibility to confirmation bias, in other words, looking for alignment with pre-existing beliefs \cite{pereira2023identity, savolainen2022what}. Additionally, individuals at ideological extremes often perceive mainstream media as hostile or biased against their perspectives \cite{gunther2017who}, which could make them less critical of alternative information sources and ``lower their guard'' towards otherwise clear evidence signals of misinformation. 

\subsection{Previous trust in the media and media consumption}

Here, a paradoxical conclusion can be drawn, as higher prior trust in news media correlated with increased trust in fake news articles. However, higher news consumption was associated with decreased trust in these same articles. This opens up an interesting avenue of analysis, where media consumption and trust in the media patterns do not seem to go hand in hand. The former can be attributed to the fact that those with high trust in news media may be less likely to perform critical evaluations of news media, making them more vulnerable to disinformation. The negative relationship between news consumption frequency and trust in fake news, on the other hand, does demonstrate the protective effect of media exposure and literacy. Regular news consumers likely develop greater familiarity with journalistic conventions and therefore a sharper critical eye.  In conclusion, while trust in media institutions may be desirable for democratic functioning \cite{vanaelst2017political}, trust ``above all'' could become a vulnerability. 

\section{Conclusion and Future Work}

We have presented and discussed the findings of a survey where we aimed at revealing factors that influenced people's perception of the media, controlling for multiple variables both at the participant level (demographics, education background or baseline levels of news consumption and political ideology) and at the content level (notably, presenting news articles written by humans and AI-generated, with and without toxic content, quotes to experts, clickbait articles, etc.). We did this over three prevalent disinformation narratives which covered topics like climate change, gender issues and COVID-19. Our results reveal key insights into the nature of trust in news media. First, the topic of news articles significantly affected trust levels, with gender-related content receiving higher trust ratings compared to climate change and vaccine narratives. Second, the stance taken toward a narrative (favourable, neutral, or contrary) interacted with the topic in complex ways, suggesting that trust strongly depends on context. Third, content-level variables such as the presence of toxic language and featuring clickbait headlines consistently reduced trust across topics, while expert quotes enhanced credibility across the board.

Perhaps most notably, our findings showed that AI-generated content was generally as trustworthy as human-written content. In some cases, particularly for content containing toxic language, AI-generated articles were actually trusted more than human-written ones, which has far-reaching implications in terms of the potential for AI technologies to be used in creating convincing disinformation.

Demographic factors also played significant roles, with age negatively correlating with trust in fake news, education showing a non-linear relationship with scepticism, and political extremity at both ends of the spectrum associated with higher susceptibility to misinformation. 

For the future, we would like to extend this analysis to more locations, consider more and more granular variables, and more importantly, test other AI systems, i.e., more modern LLMs of various sizes (in number of parameters), potentially also including multimodal generative models.

\bibliographystyle{plainnat}
\bibliography{biblio}


\end{document}